# A Context-based Disambiguation Model for Sentiment Concepts Using a Bag-of-concepts Approach

Zeinab Rajabi[1], MohammadReza Valavi[2*], Maryam Hourali[3]

[1]*Electronic and Computer Department, Malek-Ashtar University of Technology* , Email: rajabi.ze@mut.ac.ir

[2]*Associate Professor, Department of Electronic and Computer, Malek-Ashtar University of Technology* , Email: valavi@mut.ac.ir

[3]*Assistant Professor, Department of Electronic and Computer, Malek-Ashtar University of Technology,* Email: hourali@mut.ac.ir

## ABSTRACT

Introduction: With the widespread dissemination of user-generated content on different web sites, social networks, and online consumer systems such as Amazon, the quantity of opinionated information available on the Internet has been increased. Sentiment analysis of user-generated content is one of the main cognitive computing branches; hence, it has attracted the attention of many scholars in recent years. One of the main tasks of the sentiment analysis is to detect polarity within a text. The existing polarity detection methods mainly focus on keywords and their naïve frequency counts; however, they less regard the meanings and implicit dimensions of the natural concepts. Although background knowledge plays a critical role in determining the polarity of concepts, it has been disregarded in polarity detection methods.

Method: This study presents a context-based model to solve ambiguous polarity concepts using commonsense knowledge. First, a model is presented to generate a source of ambiguous sentiment concepts based on SenticNet by computing the probability distribution. Then the model uses a bag-of-concepts approach to remove ambiguities and semantic augmentation with the ConceptNet handling to overcome lost knowledge. ConceptNet is a large-scale semantic network with a large number of commonsense concepts. In this paper, the point mutual information (PMI) measure is used to select the contextual concepts having strong relationships with ambiguous concepts. The polarity of the ambiguous concepts is precisely detected using positive/negative contextual concepts and the relationship of the concepts in the semantic knowledge base. The text representation scheme is semantically enriched using Numberbatch, which is a word embedding model based on the concepts from the ConceptNet semantic network. In this regard, the cosine similarity metric is used to measure similarity and select a concept from the ConceptNet network for semantic augmentation. Pre-trained concepts vectors facilitate the more effective computation of semantic similarity among the concerned concepts.

Result: The proposed model is evaluated by applying a corpus of product reviews, called Semeval. The experimental results revealed an accuracy rate of 82.07%, representing the effectiveness of the proposed model.

## 1    Introduction

In recent years, social networks, blogs, forums, and customer review websites have provided individuals with an opportunity to express their opinions about different topics. The development of social networks has remarkably enhanced the abundance of user-generated content. In general, the data extracted from questionnaires are not of much benefit in understanding people and society; however, the data generated on social networks pave the grounds for in-depth detection of individuals and society's views. Since the social networks put people individuals in a simulated real world, where they share their views on different topics voluntarily and freely, the cognition of the data extracted from these networks is of greater value. Sentiment analysis, affective computing, and personality detection have emerged as the most popular natural language

---

processing tasks and developed the cognitive computing field in recent years [1-3]. Sentiment analysis, also known as opinion mining, refers to a field of study in which individuals' opinions, sentiments, evaluations, appraisals, attitudes, and emotions towards entities such as products, services, organizations, individuals, issues, events, topics, and their attributes are analyzed [4, 5]. The sentiment analysis mainly aims to detect the positive or negative polarity of opinions as such it has attracted the attention of many researchers over recent years.

The sentiment analysis methods have recently reached a remarkable accuracy, though they are still far away from human's ability in understanding opinions. Although there are a large number of highly accurate methods to analyze and extract the relevant knowledge from the existing structured data, the extraction of such unstructured data is still highly challenging [6]. Sentiment analysis is mainly considered as a classification problem aimed at categorizing texts or sentences into positive and negative classes. An in-depth investigation of such analyses reveals that the results are not reliable if the background knowledge is not taken into account [4],[7].

The statistical methods for the sentiment analysis blindly use keywords and count the frequency of their co-occurrence. Such methods less rely on the meaning and implicit dimensions associated with natural language concepts [8]. The concept-level approach detects an expressed sentiment more delicately and even better than the purely syntactic techniques as this technique analyzes the multiword sentiments, which do not explicitly reflect polarity and are associated with the concepts. Not benefiting from comprehensive human knowledge resources, a sentiment analysis system would hardly understand the meanings of natural language texts [9]. Therefore, accurate capture of the semantics in ambiguous words plays a critical role in facilitating the language understanding by natural language processing systems [10]. Accordingly, the present study focuses on the use of the human knowledge base to improve performance by such approaches. To this end, this study includes human background knowledge to further promote this kind of approach.

In the sentiment analysis process, it is of paramount importance to consider the semantic knowledge base. Using the commonsense knowledge representation, we present a new model for the sentiment analyses of customer product reviews. The utilization of semantic and commonsense knowledge available in the resources such as ConceptNet provides a more abstract and enriched contextual model accuracy to detect the polarity by sentiment analysis methods.

Understanding the context of discussions in social networks and considering it analyzing the users' opinions highly affect the analysis accuracy. In general, the following dimensions affect the context [11]: (1) The subjective background of individuals and the relationship among concepts in a text and concepts in individuals' minds, which are called background knowledge. Such knowledge consists of commonsense knowledge and domain-specific knowledge, and (2) The domain of discussions, for example, electronic product reviews or political issues, and (3) Concepts co-occurring with the sentiment concepts in a text.

Sentiment lexicon is the most important factor in the sentiment analysis and is commonly used for expressing positive or negative sentiments [4]. In the classical sentiment analysis, the polarity of each lexicon is considered to be independent from the domain of the document. Therefore, the existing challenge posed to the sentiment analysis is the sentiment lexicon, which may have a positive or negative polarity. Moreover, the domain of speech and the scope of dialogue significantly affect how polarity is detected. In this regard, many researchers have spared their efforts to determine the domain-dependent polarity [12], [13]; however, some issues have remained untouched. Although many studies have focused on determining polarity in a specific domain, it should be noted that the polarity is not constant even for words in a specific domain. Polarity is not only bound to domain but also depends on the context. The context of the word and the co-occurrence concepts determine polarity. Context refers to the situation in which a concept, as well as co-occurred concepts, appear. Research on ambiguous sentiment concepts has revealed the variations of polarity for a given word or concept used in a particular context [14], [15].

For example, regarding the drug domain, "weight gain" or "loss of appetite" are concepts with a positive or negative polarity based on the context. Contextual clues play a key role in determining the polarity of ambiguous concepts. To determine polarity based on the context more accurately, the following three research questions are addressed:

1. What are the ambiguous concepts?
2. What contextual clues may facilitate detecting the polarity of the ambiguous concepts?
3. How to fix the ambiguities of the ambiguous concepts using structured resources of semantic and commonsense knowledge?

In this paper, a model is presented to identify sentiment concepts characterized by ambiguous polarity, which are dependent on the context, by using statistical parameters. To this end, the ambiguous sentiment concepts are disambiguated by the contextual concepts and commonsense knowledge representations. Since the proposed model is based on the concept-level framework, this study focuses on breaking sentences down into concepts rather than words. The bag of concepts is semantically enriched by using Numberbatch, which is a word-embedding model based on the concepts from ConceptNet. Pre-trained word vectors help to compute the semantic similarity between the concepts more effectively.

The outline of this paper is as follows: Section 2 provides an overview of relevant studies. Section 3 provides more details about the proposed model. Experimental results and evaluation are presented in Section 4. The study is concluded and some suggestions for future research are put forth in the last section.

## 2    Related Work

### 2.1    Knowledge-based Sentiment Analysis

Studies using the background knowledge in the sentiment analysis are divided into two broad categories: Studies based on domain-specific knowledge (e.g., [16], [17]), and studies dealing with commonsense knowledge(e.g., [14], [15]). Some studies have adopted ontology to create knowledge from corpora semiautomatically [18] or manually [16], [17]; however, some others use publicly-available commonsense knowledge bases [14], [15], [19], [20], [21],[22]. The use of the commonsense knowledge bases is a more reasonable and applicable option since such bases are publicly available. On the other hand, there are two problems with such knowledge bases, namely data sparseness and insignificant amount of domain-specific knowledge, even though, the creation of a domain-specific knowledge base is laborious and time-consuming.

Several studies have focused on the creation of the knowledge bases for specific domains, a majority of which set out to model ontology, create knowledge base from data, and put them into use. Presenting a semi-automated method, Noferesti et al. [18], [23] developed the FactNet knowledge base using linked data to analyze the indirect opinions in the drug review domain. If we have a richer ontology-based knowledge base, sentiment analysis would be performed more precisely, and there would be less need for large corpora in the sentiment analysis.

With an emphasis on the background knowledge or ontology centralized the knowledge to extract features, many studies [16], [17], [19], [20], [21] have addressed feature/aspect level sentiment analysis and exerted the ontology to discover any connection among features. For example, an ontology was developed in [17] using ontology learning and FCA algorithm, and the domain ontology was then applied for the feature extraction. Afterward, the feature-level sentiment analysis was performed for a dataset of Twitter posts. The studies examining the product review databases has mainly focused on the feature/aspect-level sentiment analysis and feature extraction; however, the focus is different in the present study. Previous researchers ontologically discovered the relationships among a variety of features, and only a few studies [14], [15] enriched and disambiguated exploiting ontology for sentiment lexicons. These studies have dealt with the lexicon and disregarded the concept-level sentiment analysis (CLSA) framework and concept-based analysis.

More specifically, some researchers investigated implicit opinions as an objective statement implying an opinion, in which no sentiment lexicon is included [4]. Balahur et al. [24],[25] presented an automatic method to analyze the implicit opinion based on common-sense knowledge and ontology features. In their research, action chains (namely agent, action, and object) were extracted based on the semantic role labeling for this purpose. Then they created a knowledge base, called EmotiNet, for an implicit opinion analysis to 1) design an ontology containing the definitions of the significant concepts of the domain, and 2) expand the ontology by the use of available commonsense knowledge bases (ConceptNet) and other resources and then extend ontology with ConceptNet common-sense knowledge. In [26], [27], [28], Sentilo provided a formal representation of opinion sentences in the form of RDF graphs according to an ontology defining the main concepts and relationships characterizing opinion sentences. OntoSentilo is an ontology for opinion sentences, and SentiloNet is a new lexical resource facilitating the evaluation of opinions expressed by events and situations.

Domain-specific sentiment analysis studies have determined the polarity of a sentiment lexicon by using methods such as random walk algorithm [29], word sense disambiguation technique [30], feature-opinion pairs [31], label propagation [32], [33], [34], [35] [33], unsupervised machine learning by a directed acyclic graph, sentiment word, and aspect pairs [36], and fuzzy logic to model polarity [12]. Such studies disregarded context-oriented semantics, the impact of contextual clues on polarity identification, and the fact that the polarity may vary by context. In the literature, few studies have indicated that the polarity is dependent not only on the domain but also on the context, as a more remarkable factor [15],[14], [37], [38]. Saif et al. [37], [38] presented a lexicon-based approach using a contextual representation of words, called SentiCircles, which could capture the latent semantics of words from their co-occurrence patterns and update their sentiment orientations accordingly. Accordingly, they made attempts to include the context in the sentiment analysis system by using co-occurrence patterns. In many cases, the polarity is caused not only by domain but also by context, as a more important factor. Contextual knowledge contains background knowledge and co-occurrence concepts. Further attention to the co-occurrence concepts would significantly facilitate the polarity identification of implicit opinions.

As an effective method in many natural language processing tasks, deep neural networks have been accompanied with great outcomes when applied in the sentiment analysis [39, 40]. Furthermore, distributed representations or word embedding are considered as a key feature in the state-of-the-art sentiment analysis systems. These techniques encode text into the fixed-length vectors, which can be directly used by machine-learning methods and neural networks, as techniques receiving greater attention recently [41] [42]. The end-to-end deep neural networks make the system automatically learn complex features extracted from data, thereby minimizing the manual efforts. However, of these approaches is to use a large amount of annotated data [43]. Efficient training of the neural networks in small datasets is still an open challenge, the removal of which could improve the sentiment analysis systems.

An interesting approach in deep learning for sentiment analysis is to augment the knowledge included in the embedding vectors with the other information sources. Such added information can be sentiment specific word embedding (SSWE) [44] or a concatenation of manually crafted features and SSWEs [45, 46]. To be more specific, the neural sequential models such as long short-term memory (LSTM) with a capacity for representing sequential information have attracted more attention in this context. LSTM can only learn implicit knowledge from sequences by avoiding the troublesome of gradient vanishing. In contrast, background knowledge such as commonsense knowledge is hard to be learned from the data in which commonsense facts are not incorporated explicitly [22].

## 2.2   Concept-Level Sentiment Analysis

The sentiment analysis approach is classified into two broad categories [47], [48]: lexicon-based approach [49] and machine learning approach, the latter of which is subcategorized into supervised and unsupervised. Cambria [9] presented another category of concept-level sentiment analysis. The concept-level sentiment analysis emphasizes on semantic analysis of texts with regard to ontology and semantic network, thus providing an opportunity to gather emotional and conceptual information associated with different opinions. With an emphasis on a large semantic knowledge base, the approach avoids to count keywords and the number of co-occurrences blindly and instead relies on the implicit features associated with the linguistic concepts. Unlike purely syntactic methods, the concept-level approach is capable of detecting polarity expressed in a delicate manner. For example, the polarity of a sentence, which is obtained implicitly, can be discovered by concept analysis through linking concepts together.

Cambria et al., as the leading developers of such methods, introduced CLSA framework [50], a reference for those scholars who are to adopt a holistic and semantic- perspective towards sentiment analysis. The main features of this framework are as follows: (a) The main focus is on the concept-level analysis of opinionated texts rather than a word-level one, and (b) All natural language processing tasks are essential for extracting opinions from a text. CLSA framework focuses on the computational foundations of the sentiment analysis studies to determine eight key natural language processing tasks essential for the correct interpretation of an opinionated text [50]: micro text analysis, semantic parsing, subjectivity detection, anaphora resolution, sarcasm detection, topic spotting, aspect extraction, and polarity detection. In the present study, the proposed model is based on CLSA framework in which semantic parsing or concept parsing is a fundamental step.

The main difference between the traditional sentiment analysis approach and the concept-level approach is that the traditional approach relies on some parts of the text containing the sentiment lexicon which explicitly expresses the opinions; however, the relations among sentence entities and background knowledge are of great importance in the concept-level approach. In other words, the latter approach identifies the existing concepts of a text and their relations and interactions with the background knowledge and employs such relations and interactions to explore opinions conveying the sentiment implicitly.

### 2.2.1 Concept Parsing

Concept parsing is defined as the deconstruction of a natural language text into concepts [50]. A concept is an entity in a text, which can be either a simple noun, verb, adjective, adverb, or event or a complex one. In [51], simple concepts were extracted using a graph-based method. Furthermore, Agarwal et al. [52] and Poria et al. [53] proposed a set of rules for discovering complex concepts using a dependency parser. Accordingly, a concept is extracted in the form of one-word, two-words, or multi-word constructs. After extracting the concepts, the bag-of-concepts, instead of a unigram (bag-of-words), bigram, and n-gram, is available for further processing [51], which can also be used as a feed of a commonsense reasoning algorithm. In this regard, a bag-of-concepts is much better than a bag-of-words. For example, the sentence *"I contact to phone services to help me for discharging my device."* is parsed as the following: *contact_to_phone_service, phone_service, discharge_device.*

The concept extraction is one of the key stages in the concept-level sentiment analysis [50], [52]. The existing approaches extracting the concepts from a text can be classified into two categories: linguistic approach [51], [53], [52], and statistical approach [54]. The combination of the linguistic and statistical approaches improves the concept extraction process.

### 2.2.2 SenticNet

SenticNet is a sentiment lexicon resource, constructed by clustering the vector space model of affective commonsense knowledge extracted from ConceptNet in the concept-level sentiment analysis [55]. The source provides researchers with a list of concepts accompanying the polarity value and intensity in a range from -1 to +1. The polarity intensity of the presented concepts is calculated by using artificial intelligence and semantic web techniques.

In particular, SenticNet uses dimensionality reduction to calculate the affective valence of a set of Open Mind concepts and represents them in a machine-accessible and -processable format [56]. The resource contains more than 50000 single/multi- word concepts. In other words, it consists of more than 30000 multi-word concepts. For example, a query of the concept "*a lot of fun*" provides the sentiment information and a positive polarity value, as shown in Figure 1.

```
- <rdf:RDF>
  - <rdf:Description rdf:about="http://sentic.net/api/en/concept/a_lot_of_fun">
      <rdf:type rdf:resource="http://sentic.net/api/concept"/>
      <text>a lot of fun</text>
    - <semantics>
        <concept rdf:resource="http://sentic.net/api/en/concept/radical"/>
        <concept rdf:resource="http://sentic.net/api/en/concept/enjoyable"/>
        <concept rdf:resource="http://sentic.net/api/en/concept/just_fun"/>
        <concept rdf:resource="http://sentic.net/api/en/concept/good_mental_health"/>
        <concept rdf:resource="http://sentic.net/api/en/concept/fun_play"/>
      </semantics>
    - <sentics>
        <pleasantness rdf:datatype="http://w3.org/2001/XMLSchema#float">0.814</pleasantness>
        <attention rdf:datatype="http://w3.org/2001/XMLSchema#float">0</attention>
        <sensitivity rdf:datatype="http://w3.org/2001/XMLSchema#float">0</sensitivity>
        <aptitude rdf:datatype="http://w3.org/2001/XMLSchema#float">0.856</aptitude>
      </sentics>
    - <moodtags>
        <concept rdf:resource="http://sentic.net/api/en/concept/joy"/>
        <concept rdf:resource="http://sentic.net/api/en/concept/admiration"/>
      </moodtags>
    - <polarity>
        <value>positive</value>
        <intensity rdf:datatype="http://w3.org/2001/XMLSchema#float">0.557</intensity>
      </polarity>
    </rdf:Description>
```

**Figure 1: The concept "*a_lot_of_fun*" is extracted from SenticNet website demo. This concept consists of 4 words and does not exist in other sentiment lexicon resources. [57]**

### 2.2.3 ConceptNet Semantic Network

ConceptNet is a multilingual knowledge base, designed to represent words, phrases, and their common relationships that people use. It is a large semantic network consisting of a large number of commonsense concepts [58], [59], [60]. A concept is an entity defining commonsense knowledge [36]. The commonsense knowledge existing in the base is contributed and collected by ordinary Internet users around the world. It has the biggest machine-usable resource containing more than 250000 links. This knowledge base is publicly available and can be used for various text-mining inferences.

ConceptNet semantic graph represents the information of OpenMind entity in the form of a directed graph as such the nodes are concepts, and the labeled edges are commonsense assertions connecting the concepts. In other words, it consists of nodes (concepts) connected with edges (relationships among concepts). For example, the "*phone*" has the relationship of "*UsedFor*" with "*making a phone call*", and the "*cellphone*" has the relationship of "*IsATypeOf*" with the "*phone*", etc.

The relationships in the ConceptNet semantic network have been designated on the basis of the texts written by ordinary users; however, the relationships in WordNet have been defined by the experts. With a focus on the commonsense relationships among the concepts, ConceptNet aims at reasoning the commonsense knowledge. The nodes in WordNet's include purely lexical items (atomic-meaning words and simple phrases). In contrast, ConceptNet encompasses higher-order compound concepts (e.g., "talking to someone" and '" chatting with a friend") to represent knowledge for a greater range of concepts found in everyday life. Moreover, the semantic relationships in ConceptNet are extended from a triplet of synonym, "*IsA*", and "*PartOf*" to more than twenty semantic relationships such as "*EffectOf*" (causality), "*SubeventOf*" (event hierarchy), "*CapableOf*" (agent's ability),

*"MotivationOf"* (affect), *"PropertyOf"*, and *"LocationOf"*. In general, ConceptNet has an informal, undeniable, and practically valuable nature.

## 2.3    Word embedding

Previous machine learning methods generally use a one-hot vector representation, in which the dimension of the vector increases with an increase in the text data to be processed. This simply leads to dimensionality disaster and fails to capture the relationship between words and words. The semantic vector (also known as *word embedding* from a deep-learning perspective) is a technique for language modeling and feature learning, which transforms words in vocabulary to the vectors of continuous real numbers [61]. For example, the word "buy" is represented by (…, 0.25, 0.20, …). The technique normally involves embedding from a high-dimensional sparse vector space (e.g., one-hot encoding vector space, in which each word takes a dimension) into a lower-dimensional dense vector space. Each dimension of the embedding vector represents a latent feature of a word. The vectors may encode linguistic regularities and patterns. Most common approaches are train unsupervised word embedding models, which are trained based on no specific objective and mainly aim to capture language knowledge. This type of word vectors, which are trained using co-occurrence information, are normally called generic or pre-trained word vectors. Semantic vectors allow the numerical comparison of the word meanings. It can be used directly as a representation of word meanings or as a starting point for further machine learning.

Word2Vec is a commonly-used word embedding system[41], which essentially is a computationally-efficient neural network prediction model and learns word embedding from a text. It contains continuous bag-of-words (CBoW) and Skip-gram (SG) model. GloVe is another frequently-used word embedding system [42], which is trained on the nonzero entries of a global word-word co-occurrence matrix. Many deep learning models in natural language processing use word embedding results as input features [61]. *ConceptNet Numberbatch* [62] is a set of semantic vectors (word embedding) created by combining ConceptNet with the other data sources. Numberbatch vectors are measurably better for this purpose than the well-known word2vec vectors, which are trained on google news and also measurably better than GloVe vectors. This system takes word2vec and GloVe as inputs so that it can improve them. An interesting point about vector representations is that their togetherness in a set provides better results, in comparison to their being separated [63].

Some part of the information represented by these vectors comes from ConceptNet, a semantic network of knowledge about word meanings. Such embedding benefits from the fact that they have semi-structured commonsense knowledge from ConceptNet, which provides them an opportunity to learn about words not being observed in the context.

In this study, pre-trained concept vectors were used to enrich bag-of-concepts semantically and increase the model power. Unlike GloVe and word2vec vocabulary, Numberbatch supports the multi-word concept. Because of data sparseness in the ConceptNet graph, the representation of the concept as a vector of continuous real numbers remarkably contributes to measuring similarities.

## 3    Proposed Model

### 3.1    Detection of Ambiguous Sentiment Concepts

Since our model is based on the CLSA framework, the present study focuses on parsing sentences into concepts rather than words. The polarity of ambiguous sentiment concepts has a togetherness relationship with domain and changes depending on the context. Since the polarity of this category of concepts is context-dependent, context should be taken into account to achieve a more accurate polarity determination. To solve this problem, a model is proposed to detect the sentiment concepts, as described below. Figure 2 presents the proposed model in general.

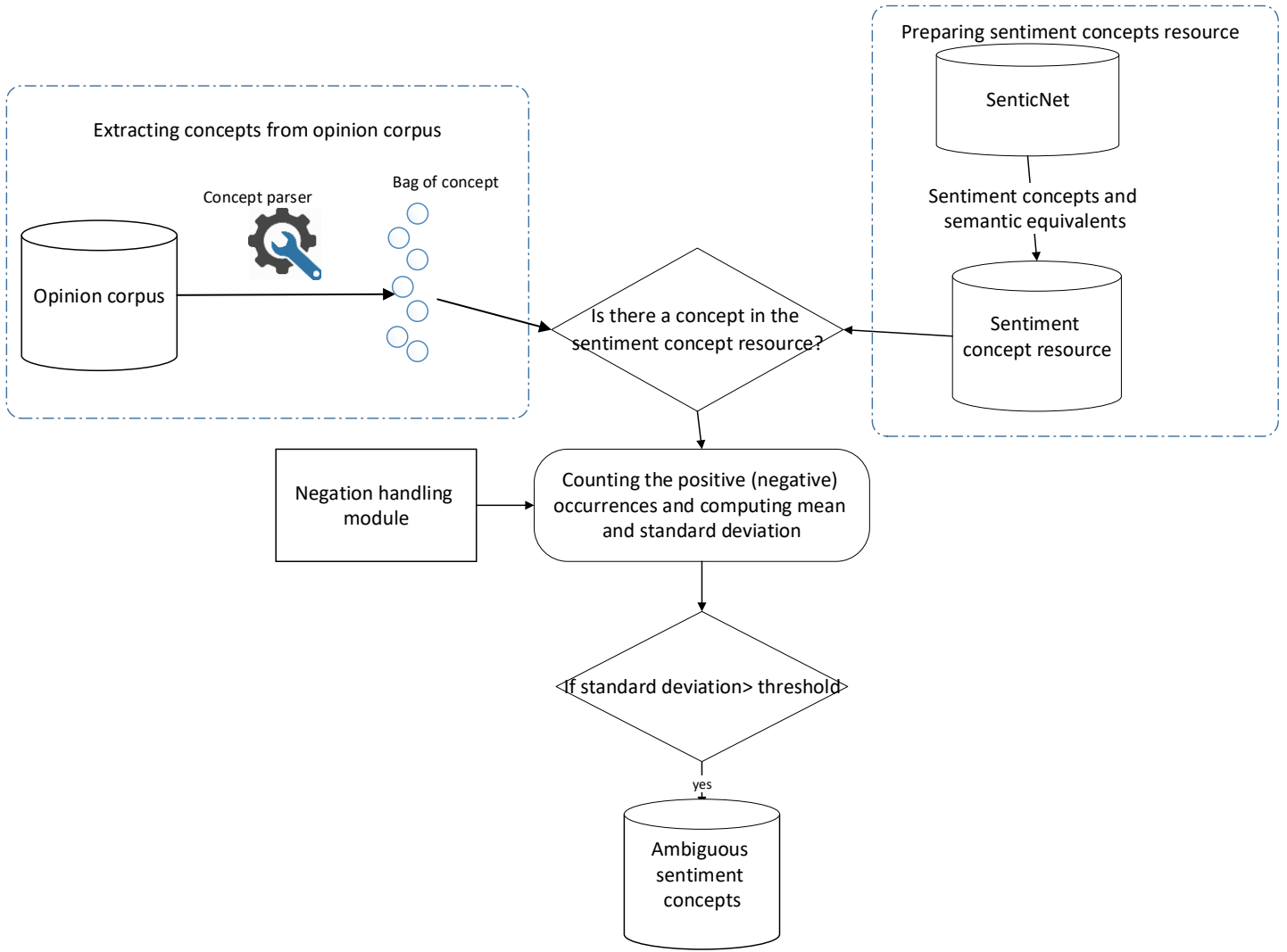

Figure 2 Model proposed to detect ambiguous sentiment concepts. The standard deviation metric separates the ambiguous concepts from the unambiguous ones.

## Step 1: Extracting Concepts from Text

In this step, the existing concepts are extracted from product review corpus using a concept parser. A concept is a single/multi-word expression (i.e., a simple noun, verb, adjective, adverb, or an event or a complex concept). Instead of using the unigram (bag-of-words), bigram, and n-gram as features in the classification, the bag-of-concepts is available for the next step of the process.

## Step 2: Preparing Resource of Sentiment Concepts

The sentiment concepts are accumulated based on the SenticNet sentiment lexicon resource. Since this resource is concept-based, it is selected from other sentiment lexicon resources such as SentiwordNet[64], MPQA[65], General Inquirer[66], and VADER[49]. Moreover, the semantic equivalents defined in the resource are added to provide wider coverage of the data. The prepared resource contains single/multi-word concepts and sentiment values. As mentioned in Section 2.2.2, the sentiment value of the resource is determined constantly; hence, it is necessary to determine polarity depending on the context of the following subsections. Finally, the sentiment concept resource $= \{sc_1, sc_2, \ldots, sc_i, \ldots, sc_n\}$ is built in this step.

## Step 3: Counting Positive and Negative Occurrences and Computing Mean and Standard Deviation

Each sentiment concept may occur in sentences with a positive or negative polarity. The proposed model uses the distribution of sentiment concepts in the corpus to determine ambiguity, in which *mean*($\mu_i$) and *standard deviation*($\delta i$) are the two critical statistical parameters, with a high *standard deviation* indicating an ambiguous concept. In this step, the occurrence of sentiment concepts in the negative and positive sentences is counted. Then the *mean* and *standard deviation* is calculated for each concept. The process of determining ambiguous concepts is described in Algorithm 1. The parameters of the equations are defined as follows: PosCount is the number of positive events, PosScore is the positive score of a sentence, NegCount is the number of negative events, and NegScore is the negative score of a sentence. Moreover, ($sc_i$) is a concept of sentiment concept resource produced in the previous step.

---

**Input:** Training corpus, Sentiment concepts resource = $\{sc_1, sc_2, \ldots, sc_i, \ldots, sc_n\}$

**Output:** Ambiguous sentiment concepts

**For** each ($sc_i$) in Sentiment concepts resource **do**

    Compute *mean*($\mu_i$) and *standard deviation*($\delta i$) with the following equations:

$$\mu_i = \frac{PosCount(sc_i)*PosScore(sc_i) - NegCount(sc_i)*NegScore(sc_i)}{PosCount(sc_i) + NegCount(sc_i)}$$

$$\delta i = \sqrt{\frac{\left(\left((\mu_i - PosScore(sc_i))^2 * PosCount(sc_i)\right) + \left((\mu_i - NegScore(sc_i))^2 * NegCount(sc_i)\right)\right)}{(PosCount(sc_i) + NegCount(sc_i))}}$$

    **If** $\delta i > 0.85$ **then** add $sc_i$ to the *Ambiguous concepts resource*

**Return** *Ambiguous concepts resource*

---

## Step 4: Separating Ambiguous Concepts from Unambiguous Ones

Standard deviation is a measure representing the amount of variation or dispersion for a set of values. A low standard deviation indicates that the values tend to be close to the mean of the set, while a high standard deviation shows that the values are spread out over a wider range. In this regard, if the standard deviation of a concept is high (higher than the *threshold*), the polarity of the concept is scattered. This means that the concept is ambiguous, and its polarity should be determined based on the context. Furthermore, if the standard deviation of a concept is low, the polarity of the concept is closed. To ensure the quality of the ambiguous concepts, we set a threshold to separate the ambiguous concepts from unambiguous ones. Human experts check the ambiguous and unambiguous list based on the context and approximately acceptable threshold. In other words, the threshold *h* is obtained empirically and according to the corpus, i.e. *0.85* in the present data. Thus, a source of sentiment concepts is obtained, which are ambiguous based on the context and need to be disambiguated and enriched in the next section.

### 3.2 Disambiguation of Sentiment Concepts

In this section, a context-oriented model of polarity determination is presented. This study does not focus on the word-sense disambiguation, which is to extract the correct meaning of polysemy words; however, this study is an attempt to improve the sentiment orientations. A context-oriented modification of ambiguous concepts necessitates paying attention to contextual clues. The contextual clues, textual or non-textual, can be used for the polarity determination of the ambiguous concepts. In this study, the emphasis is placed on the textual contextual clue, which is defined as the concepts occurring around an ambiguous concept. In addition, the semantic information latent in the commonsense knowledge base is utilized for enriching these clues. For this purpose, ConceptNet semantic network is exploited. Figure 3 shows our proposed model for the context-based disambiguation of sentiment concepts.

### 3.2.1 Collecting Contextual Concepts

Contextual concepts are those having a high frequency of co-occurrence with an ambiguous concept in the corpus. In other words, highly co-occurred concepts with an ambiguous concept are considered as contextual concepts. In this regard, two sets of co-occurred concepts are constructed for each ambiguous concept: (1) positive contextual concepts co-occurring in a positive sentence, and (2) negative contextual concepts co-occurring in a negative sentence. A contextual concept set includes the main concepts surrounding an ambiguous concept. In other words, co-occurred concepts are contextual clues of ambiguous concepts.

Extracted contextual concepts contain redundancy and noisy concepts; therefore, they should be eliminated. Agarwal et al. [52] used mRMR method to remove redundancy concepts. We use point mutual information (PMI) measures to select the contextual concepts having robust relationships with the concerned ambiguous concept. PMI measures the degree of statistical dependence between two concepts. Equation 1 shows how to compute PMI for the ambiguous concept and the contextual concept.

$$PMI(ambiguous\ concept, concept) \quad\quad (1)$$
$$= \frac{p(ambiguous\ concept, concept)}{p(ambiguous\ concept)\,p(concept)}$$

Here, $p(ambiguous\ concept, concept)$ is the actual co-occurrence probability of an ambiguous sentiment concept and a contextual concept, and $p(ambiguous\ concept)$ $p(concept)$ is the co-occurrence probability of the two concepts if they are statistically independent. Finally, a set of positive contextual concepts and a set of negative contextual concepts are created for each ambiguous sentiment concept in this section. These sets are applied for the semantic augmentation and classification in the next sections.

Context concepts$^{\pm}$ $_{ambiguous\ sentiment\ concept}$ = {$q_1^{\pm}$, $q_2^{\pm}$, …}

### 3.2.2 Semantic Augmentation Using Commonsense Knowledge

In sentiment analysis, the semantic and commonsense knowledge accessible in specific resources such as ConceptNet contributes to providing a more abstract and richer contextual modeling and reaching a better accuracy to determine the polarity. The relationships available in the external resource of the semantic knowledge base contribute to augmenting the bag-of-concepts. The semantic relationships in ConcpetNet are classified in Table 1. As it can be observed, there are 24 associated nodes for the concept "*laptop*" in the ConcpetNet, and the concept "*going on* the internet" has the relationship of "*Things that require*" with the concept "laptop".

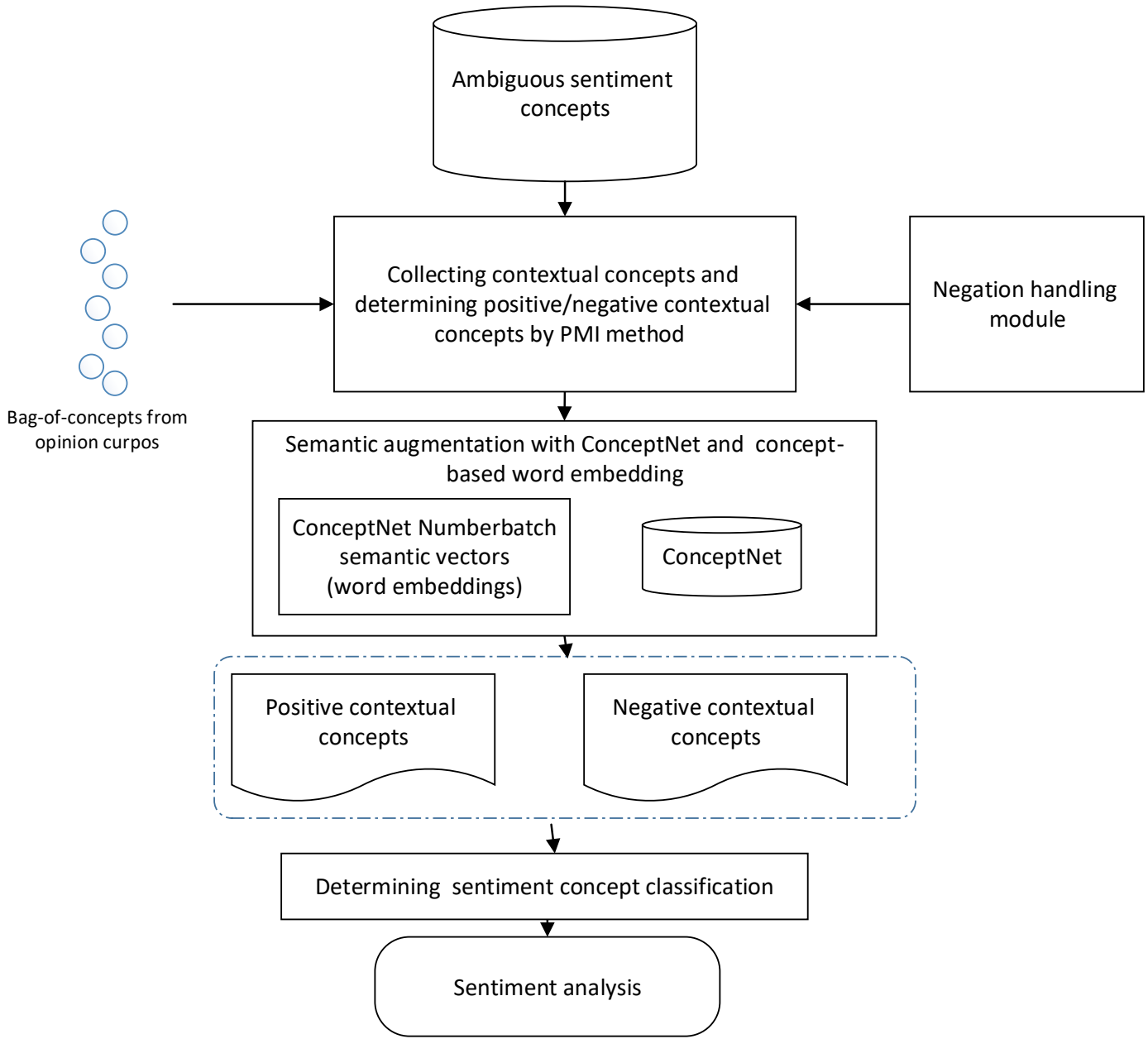

Figure 3: Model proposed for disambiguating sentiment concepts. Two sets of collected concepts (positive contextual concepts/ negative contextual concepts) are constructed for each ambiguous concept. They are applied in a sentiment analysis system for sentiment classification.

In order to augment the contextual concepts semantically, the candidate concepts are extracted from ConceptNet semantics network. The candidate concepts are the neighbors of ambiguous concepts, which are directly connected to the ambiguous sentiment concept in ConceptNet semantic network by an edge. They have the potential to be added to the bag-of-concepts through further examination. Figure 4 shows

*the* candidate concepts = $\{o_1, o_2, \ldots, o_i, \ldots, o_n\}$ and their relationships with the ambiguous sentiment concept in the ConceptNet semantic network.

ConceptNet Numberbatch is a word embedding model, which is also known as a set of semantic vectors. It can be used directly as a representation of word meanings or as a starting point for further natural language processing. Numberbatch is built using an ensemble combining data from ConceptNet, word2vec, and GloVe using a variation on retrofitting. This resource has a vector of real numbers for each concept from ConceptNet. Numerical vectors are used to better select the concepts and add them to the bag-of-concepts.

Table 1: Main categories of relationships in ConceptNet

| Things | IsA, Part-Of, MemberOf, HasA, HasProperty, Synonym, Antonym, DerivedFrom, DefinedAs, TranslationOf, SimilarTo |
|---|---|
| Functional | UsedFor, CapableOf |
| Spatial | AtLocation, LocatedNear |
| Event | HasSubevent, HasFirstSubevent, HasLastSubevent, HasPrerequisite |
| Causal | Causes, Desires |
| Motivation | MotivatedByGoal, ObstructedBy |
| Other | RelatedTo, CreatedBy, MadeOf |

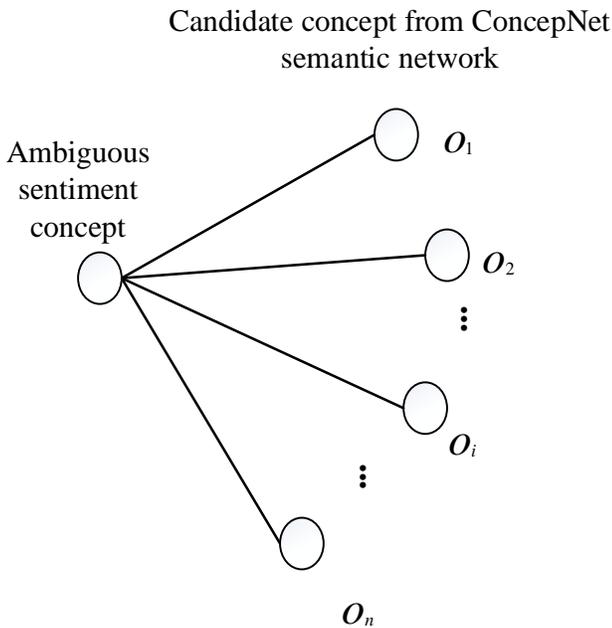

Candidate concept from ConcepNet semantic network

Figure 4: Candidate concepts from CocnepNet semantic network

To define the similarity between two target words $V$ and $W$, natural language processing uses a measure for considering two of such vectors and providing a measure of vector similarity.

Thus far, the most common similarity metric is the cosine of the angle between the vectors. Cosine similarity is a measure of similarity between two non-zero vectors of a dot product space, which measures the cosine of the angle between these vectors. Equation 2 shows the normalized dot product of two vectors.

$$\cos(\vec{V} . \vec{W}) = \frac{\vec{V} \bullet \vec{W}}{|\vec{V}||\vec{W}|} = \frac{\sum_{i=1}^{N} V_i W_i}{\sqrt{\sum_{i=1}^{N} V_i} \sqrt{\sum_{i=1}^{N} W_i}} \qquad 2$$

Cambria et al. [67] applied the cosine similarity measure for a relevant lexical substitute from ConceptNet network. In this study, this metric was used to measure similarity and choose a concept from ConceptNet for semantic augmentation. Our solution is based on the assumption that the addition of a relevant concept should semantically be similar to both the sentiment concept and the contextual concepts. In the ConceptNet Numberbatch, one vector is presented for each concept. Cosine similarity is applied for ConceptNet Numberbatch embedding vectors.

To select a concept from the concepts having a direct link with the ambiguous concept, this study applies cosine similarity metrics. Equation 3 illustrates the use of the cosine similarity measure to compute the similarity between the vector representation of the candidate concept ($o_i$) and that of an ambiguous sentiment concept as well as the similarity between candidate concept ($o_i$) and contextual concept ($q_j$). Moreover, $q_j$ belongs to the union of two sets of positive/negative contextual concepts. The detail of the contextual concept extraction is described in Section 3.2.1.

The similarity value is computed for each candidate concept ($o_i$) in the ConceptNet. Some candidate concepts with high similarity values are added to the contextual concept sets (bag-of-concepts). This step leads to the extension of each one/multi-word concept extracted from the sentences in the previous section. The bag-of-concepts is thus extended and semantically enriched. For example, "*going to market*" has "*Subevents of*" relationship with "*buy*", so that "*going to market*" expands to the concept "*buy*" and is added to the bag-of-concepts from the concerned sentence.

$$Similarity(o_i) = \cos(amg, \ o_i) + \sum_{j=1}^{m} \cos(q_j, o_i) \qquad 3$$

### 3.2.3 Handling Negation

In a sentiment analysis system, negation plays a key role in polarity detection and shifts the sentiment value. In other words, it may turn a positive polarity into a negative one and vice versa. Hence, it is necessary to recognize and apply negation in any sentiment analysis system. The concept-level sentiment analysis approach requires negation recognition. In this paper, negation markers of the sentences are detected by the method of listing common negation keywords [32], [68] and are used in the disambiguation process.

When collecting the contextual concepts of the positive sentences from the corpus, they are considered if the negation clues exist in the sentence, and the contextual concept is included in the negative set instead of the positive set. Moreover, the contextual concept in negative sentences is included in the positive set if the negation clues exist in the sentence. In other words, the polarity of the contextual concept is reversed if a negation clue is recognized in a sentence.

**If** (*score(sentence)>0* and *negation clue is present*) **Then put** *contextual concept* **in the** *negative set*

**If** (*score(sentence)<0* and *negation clue is present*) **Then put** *contextual concept* **in the** *positive set*

This study applies negation markers extracted from VADER-Sentiment-Analysis package[49].

### 3.2.4 Determining Sentiment Concept Classification

In Section 3.2.1, two sets of collected concepts (positive contextual concepts/ negative contextual concepts) are created for each ambiguous concept. First, positive contextual concepts keep positive contextual clues. Second, the negative contextual concepts keep negative contextual clues. Each set is a small bag-of-concepts, which is semantically extended using Numberbatch, as mentioned in Section 3.2.2. The two sets are stored beside the ambiguous concept and are used when it is necessary to determine the polarity of the ambiguous concept.

After creating two sets, semantic augmentation, and negation handling, we employ the Naïve Bayes technique to determine the precise polarity. The Naive Bayes technique approximates the polarity of an ambiguous concept based on the probabilities of the collected positive/negative contextual concepts. Positive/negative features are concepts that have been selected and augmented.

The probability for a concept C, denoted by P ($C_{+/-}$), to be positive or negative is determined according to the probability of contextual clues (C = {$c_1$, $c_2$, .. $c_n$}). To determine the polarity of an ambiguous concept, a class (positive or negative) with the *highest* probability needs to be selected. Equations 4, 5, and 6 show how to determine the polarity of an ambiguous concept.

$$c = \{c_1, c_2, \ ... \ , c_i, ..., c_n\} \qquad (4)$$

$$p(C^-/c) = \frac{p(C^-) \prod_{i=1}^{n} p(c_i/C^-)}{\prod_{i=1}^{n} p(ci)} \qquad (5)$$

$$p(C^+/c) = \frac{p(C^+) \prod_{i=1}^{n} p(c_i/C^+)}{\prod_{i=1}^{n} p(c_i)} \qquad (6)$$

## 4 Experimental Study

### 4.1 Pre-processing

To verify the proposed model, we conduct experiments on an electronic product review corpus, called SemEval2015. The corpus is retrieved from the GitHub website. The corpus used by researchers to evaluate the sentiment analysis research was standardized and published on the Internet. The corpus contains 5545 sentences written by Internet users for reviewing electronic products. The sentences included in the corpus are annotated with a positive, negative, or neutral sentiment.

Pre-processing the data is conducted as follows: Correcting the punctuations and words to have a superior sentence distinction, eliminating the duplicate letters of words, correcting the misspelled words, parentheses, and symbols "?!", replacing the abbreviations, and converting the complex sentences into simple ones. Moreover, the name entities are recognized and substituted. A major part of the data processing, including sentence tokenization, word tokenization, and lemmatization, was performed using NLTK(http://nltk.org) packages. Then the concepts in each sentence are extracted by a concept parser[2] developed by the Sentic group. SenticNet resource [3] and semantic equivalents are utilized to create the sentiment concept resource. The connections between the concepts of ConceptNet semantic network are provided by linked data, and JSON format and Lookup () function are also used to navigate the semantic network.

### 4.2    Results

The performance of the proposed model is evaluated using *precision*, *recall*, *f-measure*, and *accuracy* measures. Equations 7, 8, 9, and 10 show the computation of these measures based on *true positive (TP)*, *true negative (TN)*, *false positive (FP)*, and *false negative (FN)* values.

$$Precision = \frac{TP}{TP + FP} \qquad\qquad 7$$

$$Recall = \frac{TP}{TP + FN} \qquad\qquad 8$$

$$F1 - measure = \frac{2 * Recall * Precision}{Recall + Precision} \qquad 9$$

$$Accuracy = \frac{TP + TN}{TP + TN + FP + FN} \qquad 10$$

We performed a 5-fold cross-validation strategy to divide the sample set into five parts, in which four parts were a training dataset, and the other part was a verification dataset. Then we repeated this process five times and used the mean as the final result. Table 3 reports the mean performance in terms of classification accuracy.

To evaluate the performance of this study, we used the hybrid approach (lexicon-based and supervised machine learning) [48] for the sentiment analysis system. The bag-of-concepts as a context enriched by common-sense knowledge is considered as a feature in the classification. Pang et al. [69] exploited a bag-of-words as a feature in Bayesian classifier and reached 81.0 accuracy in a movie review dataset for the two positive and negative categories. In this paper, we use their model for electronic product reviews and compare it with the proposed model. In this regard, the bag-of-words and bag-of-concepts from the text representation schemes are also tested to compare the proposed model with the baseline. The disambiguated sentiment lexicons with unambiguous sentiment lexicons are used to empower the features. Naïve Bayes and SVM classifiers that achieved successful results in previous works [47]are also used for the concerned classification.

The highest *f-measure=84.53%* and *accuracy= 82.07%* were obtained using Naïve Bayes classifier and bag-of-concepts as a context enriched by common-sense knowledge feature. Comparing with the method that is based on the bag-of-words, these two methods demonstrate a significant improvement. The results of the evaluations and comparisons are presented in Table 3.

Table 2 exemplifies a couple of sentences, their concepts specified by the concept parser, and the detected ambiguous contextual concepts. For example, "*Save your money and go for a better device*". The sentence is parsed into the following concepts: *"go_money"*, *"save_money"*, *"go_for_device"*, *"better_device"*, *"device"*, *"save"*. It seems that "*better device*" is positive and makes the sentence be positive; however, it is an ambiguous concept. The concept *"save_money"* is a negative contextual concept for *"better_device"* and it leads to the detection of negative polarity.

### 4.3    Discussion

As previously mentioned, the knowledge-based sentiment analysis approach is applied to the external knowledge to increase the performance of a sentiment analysis system. It is to use pre-prepared knowledge for the sentiment analysis. On the other hand, the machine-learning approach disregards the semantic concepts of sentence and background connections and emphasizes more on the keywords and their co-occurrences. The proposed model integrates machine-learning and knowledge-based approaches and takes benefits from both approaches.

---

[2] https://github.com/SenticNet/concept-parser

[3] https://sentic.net/downloads/

Available sentiment resources are static, indicating that each term is assigned a fixed polarity. In a majority of cases, the polarity determination requires the consideration of contextual clues. Instead of putting the bag-of-words, bigram, and n-gram into practice, the proposed model uses concepts and relationships among sentence elements and semantically augments the concepts utilizing some external knowledge resources. In other words, the model considers the sentence concepts as the clues of the context and employs them to modify the ambiguity of such concepts. This involves high-frequency concepts surrounding ambiguous concepts, which play a key role in context recognition.

In this paper, the following four limitations need to be addressed: low accuracy of concept parser, negation handling, learning Numberbatch sentiment-specific word embedding to analyze emotions and covering information in the text with external knowledge resources.

1) Accuracy of concept parser: The accuracy of the proposed model is highly dependent on the accuracy of the concept parser. The more accurate the concept parsing of a sentence with fewer errors is, the better the results of the model are.

2) Learning Numberbatch sentiment-specific word embedding: The most serious problem with the word embedding learning algorithms is that they only model the contexts of words while ignoring the sentiment information [44, 46]. Learning Numberbatch word embedding in combination with the sentiment analysis task may result in better performance.

3) Negation handling problem: In the sentiment analysis system, negation is a highly important linguistic expression since it affects the polarity of the other words. The accuracy of the proposed model depends on the negation recognition and negation handling so that the better the negation recognition is, the higher the accuracy of the model will be.

4) Covering information included in the text using the external knowledge resources: The study focuses on commonsense concepts used by the users talking about buying. The broad overlap of the concepts extracted from text corpus with knowledge resources such as conceptNet and sentiment lexicon resources such as senticNet has a significant impact on achieving better results.

**Table 2: Examples of disambiguating of ambiguous sentiment concepts**

| Sentence | Polarity of sentence | Bag-of-concepts | Ambiguous sentiment concept | Contextual concept |
|---|---|---|---|---|
| Save your money and go for a better device. | negative | go_money, save_money, go_for_device, better_device, device, save | better_device (positive) | save_money (negative contextual concept) |
| I doubt you'd be overly disappointed if you invest in this machine. | positive | invest_in_machine, machine, doubt, overly, disappoint, invest | disappointed (negative) | Doubt (positive contextual concept) |
| It was worth the few ($100) extra dollars. | positive | be_worth_dollar, few_dollar, extra_dollar, dollar, worth, be_worth | extra dollar (negative) | Worth (positive contextual concept) |
| I still expected better product quality for this price range. | negative | expect_product_quality price_range better_quality product_quality, expect | better_quality (positive) | Expected (negative contextual concept) |

**Table 3: Results of polarity classification based on proposed model and baseline**

| | | Precision | Recall | F-Measure | Accuracy |
|---|---|---|---|---|---|
| **NB** | Bag-of-words (baseline) | 78.43 | 80.19 | 71.19 | 79.86 |
| | Bag-of-concepts as context | 78.18 | 85.03 | 74.61 | 77.11 |
| | Bag-of-concepts as context and common-sense knowledge (proposed model) | 80.24 | 89.32 | **84.53** | **82.07** |
| **SVM** | Bag-of-words (baseline) | 78.43 | 85.19 | 71.19 | 78.45 |
| | Bag-of-concepts as context | 75.08 | 79.1 | 74.61 | 79.51 |
| | Bag-of-concepts as context and common-sense knowledge (proposed model) | 74.42 | 90 | 81.47 | 76.07 |

## 5    Conclusion

The current study proposed a polarity disambiguation model using a bag-of-concepts and commonsense knowledge. The research was an attempt to identify ambiguous concepts and modify them by employing CLSA framework. The bag-of-concepts makes it possible to decompose a text more deeply. The involvement of the background knowledge, especially commonsense knowledge, in the sentiment analysis process can increase accuracy. In this study, the semantic knowledge provided in ConceptNet and their word embedding were used to represent commonsense knowledge. Extracting concepts and understanding their connections are the foreparts of analyzing implicit opinions, which provide a deeper understanding of a text.

The ambiguous concepts were identified using training corpus and statistical methods, and a concept-level resource, called SenticNet, was considered as the base. Contextual concepts were augmented by linking the concepts together based on commonsense knowledge. The sentiment concepts were also contextualized and disambiguated using the Naive Bayesian technique. The experimental results indicated that our proposed model achieved relatively acceptable accuracy.

When there is no enough annotated data available for training, the deep learning methods may not be appropriate. However, the embedding layer of deep learning can be used with traditional classifier methods such as Naïve Bayes and SVM to reach satisfactory results in improving the sentiment analysis prediction. In sum, this work takes advantage of a set of existing traditional sentiment classifiers as well as the combination of external knowledge resource and generic word embedding.

According to the findings mentioned above, it can be concluded that the proposed model worked well in classifying the polarities of the product reviews. It is thus suggested to extend this model in future studies to analyze the implicit and indirect opinions in a more focused way. In addition, domain-specific knowledge can be used as a complement to commonsense knowledge in order to improve the proposed model and achieve better results. Moreover, it is also recommended to have further studies on methods that combine the information available in text corpora with the external knowledge resources in the deep neural network. Such knowledge can be leveraged to include additional information not available in text corpora to promote semantic coherence.